# Stiffness Analysis of 3-d.o.f. Overconstrained Translational Parallel Manipulators

Anatoly Pashkevich, Damien Chablat, Philippe Wenger

*Abstract*— **The paper presents a new stiffness modelling method for overconstrained parallel manipulators, which is applied to 3-d.o.f. translational mechanisms. It is based on a multidimensional lumped-parameter model that replaces the link flexibility by localized 6-d.o.f. virtual springs. In contrast to other works, the method includes a FEA-based link stiffness evaluation and employs a new solution strategy of the kinetostatic equations, which allows computing the stiffness matrix for the overconstrained architectures and for the singular manipulator postures. The advantages of the developed technique are confirmed by application examples, which deal with comparative stiffness analysis of two translational parallel manipulators.**

## I. Introduction / Related works

Relative to serial manipulators, parallel manipulators are claimed to offer an improved stiffness-to-mass ratio and better accuracy. This feature makes them attractive for innovative machine-tool structures for high speed machining [1, 2, 3]. When a parallel manipulator is used as a Parallel Kinematic Machine (PKM), stiffness becomes a very important issue in its design [4, 5, 6, 7]. This paper presents a general method to compute the stiffness analysis of 3-dof overconstrained translational parallel manipulators.

Generally, the stiffness analysis of parallel manipulators is based on a kinetostatic modeling [8], which proposes a map of the stiffness by taking into account the compliance of the actuated joints. However, this method is not appropriate for PKM whose legs are subject to bending [9].

Several methods exist for the computation of the stiffness matrix: the Finite Element Analysis (FEA) [10], the matrix structural analysis (SMA) [11], and the virtual joint method (VJM) that is often called the lumped modeling [8].

The FEA is proved to be the most accurate and reliable, however it is usually applied at the final design stage because of the high computational expenses required for the repeated re-meshing of the complicated 3D structure over the whole workspace. The SMA also incorporates the main ideas of the FEA, but operates with rather large elements – 3D flexible beams describing the manipulator structure. This leads obviously to the reduction of the computational expenses, but does not provide clear physical relations required for the parametric stiffness analysis. And finally, the VJM method is based on the expansion of the traditional rigid model by adding the virtual joints (localized springs), which describe the elastic deformations of the links. The VJM technique is widely used at the pre-design stage.

Next section introduces a general methodology to derive the kinematic and stiffness model. Section 3 describes the manipulator compliant elements and the link stiffness evaluation methods. Finally in section 4, we apply our method on two application examples.

A. Pashkevich is with the IRCCyN (UMR CNRS 6597), Nantes, France and with the Department of Automatics and Production Systems, École des Mines de Nantes, France (anatol.pashkevich@emn.fr);

D. Chablat is with the IRCCyN (UMR CNRS 6597), Nantes, France (Damien.Chablat@irccyn-nantes.fr);

P. Wenger is with the IRCCyN (UMR CNRS 6597), Nantes, France (Philippe.Wenger@irccyn-nantes.fr).

## II. General Methodology

### A. Manipulator Architecture

Let us consider a general 3 d.o.f. translational parallel manipulator, which consists of a mobile platform connected to a fixed base by three identical kinematics chains (Fig. 1). Each chain includes an actuated joint "Ac" (prismatic or rotational) followed by a "Foot" and a "Leg" with a number of passive joints "Ps" inside. Certain geometrical conditions are assumed to be satisfied with respect to the passive joints to eliminate the platform rotations and to achieve stability of its translational motions.

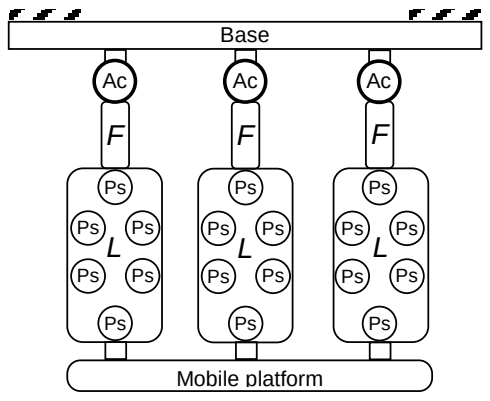

Fig. 1. Schematic diagram of a general 3-d.o.f. translational parallel manipulator (Ac – actuated joint, Ps – passive joints, F – foot, L - Leg)

Typical examples of such architectures are:

(a) 3-PUU translational PKM (Fig 2a); where each leg consists of a rod ended by two U-joints (with parallel intermediate and exterior axes), and active joint is driven by linear actuator [13];

(b) Delta parallel robot (Fig 2b) that is based on the 3-RRPaR architecture with parallelogram-type legs and rotational active joints [14];

(c) Orthoglide parallel robot (Fig 2c) that implements the 3-PRPaR architecture with parallelogram-type legs and translational active joints [10].

Here R, P, U and Pa denote the revolute, prismatic, universal and parallelogram joints, respectively.

It should be noted that examples (b) and (c) illustrate overconstrained mechanisms, where some kinematic constrains are redundant but do not affect the resulting degrees of freedom. However, most of the past works deal with non-overconstrained architectures, which motivates the subject of this paper [8].

### B. Basic Assumptions

To evaluate the manipulator stiffness, let us apply a modification of the virtual joint method (VJM), which is based on the lump modeling approach [8, 10]. According to this approach,

(a) 3-PUU translational PKM [13] (b) Delta parallel robot [14] (c) Orthoglide parallel robot [10]

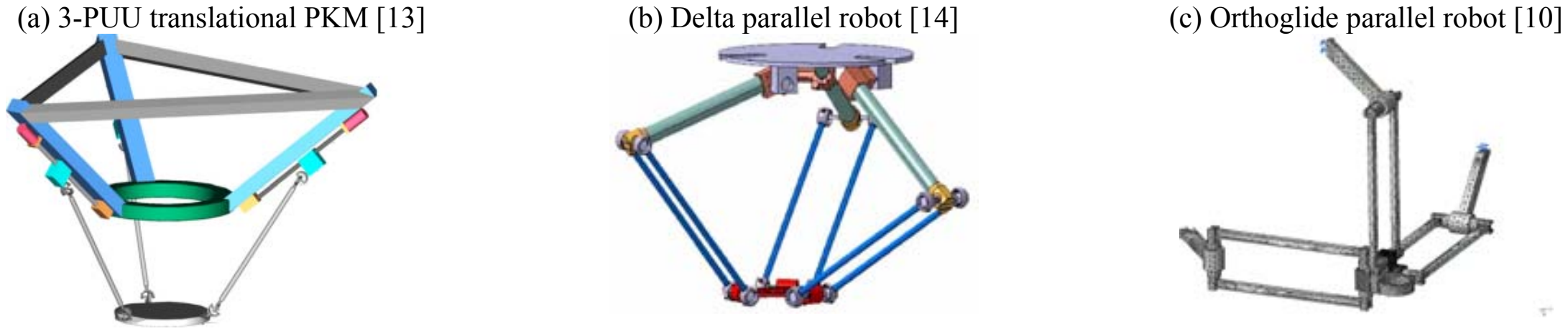

Fig. 2. Typical 3 d.o.f. translational parallel mechanisms

the original rigid model should be extended by adding the virtual joints (localized springs), which describe elastic deformations of the links. Besides, virtual springs are included in the actuating joints to take into account stiffness of the control loop. To overcome difficulties with parallelogram modeling, let us first replace the manipulator legs (see Fig. 3) by rigid links with configuration-dependent stiffness.

This transforms the general architecture into the extended 3-xUU case allowing treating all the considered manipulators in the similar manner. Under such assumptions, each kinematic chain of the manipulator can be described by a serial structure (Fig. 3), which includes sequentially:

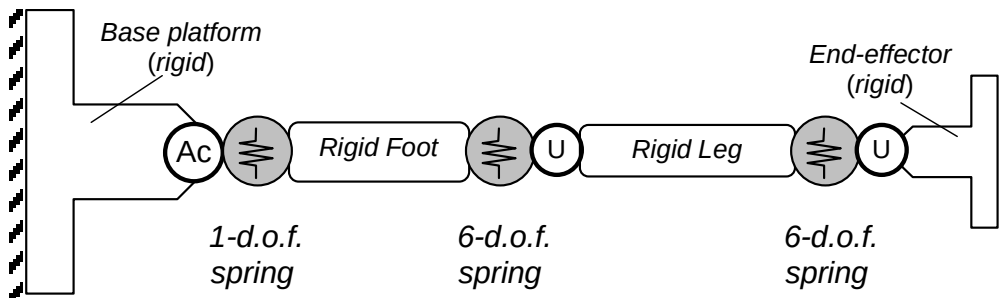

Fig. 3. Flexible model of a single kinematic chain

(a) a rigid link between the manipulator base and the *i*th actuating joint (part of the base platform) described by the constant homogenous transformation matrix $\mathbf{T}^i_{base}$ ;

(b) a 1-d.o.f. actuating joint with supplementary virtual spring, which is described by the homogenous matrix function $\mathbf{V}_a(q_0^i+\theta_0^i)$ where $q_0^i$ is the actuated coordinate and $\theta_0^i$ is the virtual spring coordinate;

(c) a rigid "Foot" linking the actuating joint and the leg, which is described by the constant homogenous transformation matrix $\mathbf{T}_{foot}$ ;

(d) a 6-d.o.f. virtual joint defining three translational and three rotational foot-springs, which are described by the homogenous matrix function $\mathbf{V}_s(\theta_1^i,\ldots\theta_6^i)$ , where $\{\theta_1^i,\theta_2^i,\theta_3^i\}$ and $\{\theta_4^i,\theta_5^i,\theta_6^i\}$ correspond to the elementary translations and rotations respectively;

(e) a 2-d.o.f. passive U-joint at the beginning of the leg allowing two independent rotations with angles $\{q_1^i,q_2^i\}$ , which is described by the homogenous matrix function $\mathbf{V}_{u1}(q_1^i,q_2^i)$ ;

(f) a rigid "Leg" linking the foot to the movable platform, which is described by the constant homogenous matrix transformation $\mathbf{T}_{leg}$ ;

(g) a 6-d.o.f. virtual joint defining three translational and three rotational leg-springs, which are described by the homogenous matrix function $\mathbf{V}_s(\theta_7^i,\ldots\theta_{12}^i)$ , where $\{\theta_7^i,\theta_8^i,\theta_9^i\}$ and $\{\theta_{10}^i,\theta_{12}^i,\theta_{12}^i\}$ correspond to the elementary translations and rotations, respectively;

(h) a 2-d.o.f. passive U-joint at the end of the leg allowing two independent rotations with angles $\{q_3^i,q_4^i\}$ , which is described by the homogenous matrix function $\mathbf{V}_{u2}(q_3^i,q_4^i)$ ;

(i) a rigid link from the manipulator leg the end-effector (part of the movable platform) described by the constant homogenous matrix transformation $\mathbf{T}^i_{tool}$ ;

The expression defining the end-effector location subject to variations of all coordinates of a single kinematic chain may be written as follows

$$\begin{aligned}\mathbf{T}_i = \mathbf{T}^i_{base}\cdot\mathbf{V}_a(q_0^i+\theta_0^i)\cdot\mathbf{T}_{foot}\cdot\mathbf{V}_s(\theta_1^i,\ldots\theta_6^i)\cdot\\ \cdot\mathbf{V}_{u1}(q_1^i,q_2^i)\cdot\mathbf{T}_{leg}\cdot\mathbf{V}_s(\theta_7^i,\ldots\theta_{12}^i)\cdot\mathbf{V}_{u2}(q_3^i,q_4^i)\cdot\mathbf{T}^i_{tool}\end{aligned} \qquad (1)$$

where matrix function $\mathbf{V}_a(.)$ is either an elementary rotation or translation, matrix functions $\mathbf{V}_{u1}(.)$ and $\mathbf{V}_{u2}(.)$ are compositions of two successive rotations, and the spring matrix $\mathbf{V}_s(.)$ is composed of six elementary transformations. In the rigid case, the virtual joint coordinates $\theta_0^i,\ldots\theta_{12}^i$ are equal to zero, while the remaining ones (both active $q_0^i$ and passive $q_0^i,\ldots q_4^i$ ) are obtained through the inverse kinematics, ensuring that all three matrices $\mathbf{T}_i$ , $i=1,2,3$ are equal to the prescribed one that characterizes the spatial location of the moving platform (kinematic loop-closure equations). Particular expressions for all components of the product (1) may be easily derived using standard techniques for the homogenous transformation matrices. It should be noted that the kinematic model (1) includes 18 variables (1 for active joint, 4 for passive joints, and 13 for virtual springs). However, some of the virtual springs are redundant, since they are compensated by corresponding passive joints with aligning axes or by combination of passive joints. For computational convenience, nevertheless, it is not reasonable to detect and analytically eliminate redundant variables at this step, because the developed below technique allows easy and efficient computational elimination.

## C. *Differential Kinematic Model*

To evaluate the manipulator ability to respond to the external forces and torques, let us first derive the differential equation describing relations between the end-effector location and small variations of the joint variables. For each *i*th kinematic chain, this equation can be generalized as follows

$$\delta\mathbf{t}_i = \mathbf{J}^i_\theta\cdot\delta\boldsymbol{\theta}_i + \mathbf{J}^i_q\cdot\delta\mathbf{q}_i,\quad i=1,2,3, \qquad (2)$$

where the vector $\delta\mathbf{t}_i=(\delta p_{xi},\delta p_{yi},\delta p_{zi},\delta\varphi_{xi},\delta\varphi_{yi},\delta\varphi_{zi})^T$ describes the translation $\delta\mathbf{p}_i=(\delta p_{xi},\delta p_{yi},\delta p_{zi})^T$ and the rotation $\delta\boldsymbol{\varphi}_i=(\delta\varphi_{xi},\delta\varphi_{yi},\delta\varphi_{zi})^T$ of the end-effector with respect to the Cartesian axes; vector $\delta\boldsymbol{\theta}_i=(\delta\theta_0^i,\ \ldots\delta\theta_{12}^i\ )^T$ collects all virtual joint coordinates, vector $\delta\mathbf{q}_i=(\delta q_1^i,\ldots\delta q_4^i\ )^T$ includes all passive joint coordinates, symbol '$\delta$' stands for the variation with respect to the rigid case values, and $\mathbf{J}^i_\theta$ , $\mathbf{J}^i_q$ are the matrices of sizes 6×13 and 6×4 respectively. It should be noted that the derivative for the actuated coordinate $q_0^i$ is not included in $\mathbf{J}^i_q$ but it is represented in the first column of $\mathbf{J}^i_\theta$ through variable $\theta_0^i$. The desired matrices $\mathbf{J}^i_\theta$ , $\mathbf{J}^i_q$ , which are the only parameters of the differential model (2), may be computed from (1) analytically, using some software support tools, such as Maple, MathCAD or Mathematica. However, a straightforward differentiation usually yields very awkward expressions that are not convenient for further computations. On the other hand, the fractionized structure of (1), where all variables are separated, allows applying an efficient semi-analytical method. To present this technique, let us assume that for the particular virtual joint variable $\theta_0^i$ the model (1) is rewritten as

$$\mathbf{T}_i = \mathbf{H}_{ij}^1 \cdot \mathbf{V}_{\theta j}(\theta_j^i) \cdot \mathbf{H}_{ij}^2, \qquad (3)$$

where the first and the third multipliers are the constant homogenous matrices, and the second multiplier is the elementary translation or rotation. Then the partial derivative of the homogenous matrix $\mathbf{T}_i$ for the variable $\theta_j^i$ at point $\theta_j^i = 0$ may be computed from a similar product where the internal term is replaced by $\mathbf{V}'_{\theta j}(.)$ that admits very simple analytical presentation. In particular, for the elementary translations and rotations about the X-axis, these derivatives are:

$$\mathbf{V}'_{Tran_x} = \begin{bmatrix} 0 & 0 & 0 & 1 \\ 0 & 0 & 0 & 0 \\ 0 & 0 & 0 & 0 \\ 0 & 0 & 0 & 0 \end{bmatrix}; \quad \mathbf{V}_{Rot_x} = \begin{bmatrix} 0 & 0 & 0 & 0 \\ 0 & 0 & 1 & 0 \\ 0 & -1 & 0 & 0 \\ 0 & 0 & 0 & 0 \end{bmatrix}. \qquad (4)$$

Furthermore, since the derivative of the homogenous matrix $\mathbf{T}'_i = \mathbf{H}_{ij}^1 \cdot \mathbf{V}'_{\theta j}(\theta_j^i) \cdot \mathbf{H}_{ij}^2$ may be presented as

$$\mathbf{T}'_i = \begin{bmatrix} 0 & \varphi'_{iz} & -\varphi'_{iy} & p'_{ix} \\ -\varphi'_{iz} & 0 & \varphi'_{ix} & p'_{iy} \\ \varphi'_{iy} & -\varphi'_{ix} & 0 & p'_{iz} \\ 0 & 0 & 0 & 0 \end{bmatrix}, \qquad (5)$$

then the desired *j*th column of $\mathbf{J}_0^i$ can be extracted from $\mathbf{T}'_i$ (using the matrix elements $T'_{14}$, $T'_{24}$, $T'_{34}$, $T'_{23}$, $T'_{31}$, $T'_{12}$ ).

The Jacobians $\mathbf{J}_q^i$ can be computed in a similar manner, but the derivatives are evaluated in the neighborhood of the "nominal" values of the passive joint coordinates $q_{j\,nom}^i$ corresponding to the rigid case (these values are provided by the inverse kinematics). However, simple transformation $q_j^i = q_{j\,nom}^i + \delta q_j^i$ and corresponding factoring of the function $\mathbf{V}_{qj}(q_j^i) = \mathbf{V}_{qj}(q_{j\,nom}^i)\mathbf{V}_{qj}(\delta q_j^i)$ allow applying the above approach. It is also worth mentioning that this technique may be used in analytical computations, allowing one to avoid bulky transformations produced by the straightforward differentiating.

### *D. Kinetostatic and Stiffness Models*

For the manipulator kinetostatic model, which describes the force-and-motion relation, it is necessary to introduce additional equations that define the virtual joint reactions to the corresponding spring deformations. In accordance with the adopted stiffness model, three types of virtual springs are included in each kinematic chain:

- 1-d.o.f. virtual spring describing the actuator compliance;
- 6-d.o.f. virtual spring describing compliance of the foot;
- 6-d.o.f. virtual spring describing compliance of the leg.

Assuming that the spring deformations are small enough, the required relations may be expressed by linear equations

$$\left[\tau_{\theta 0}^i\right] = K_{act}\left[\theta_0^i\right]; \quad \begin{bmatrix} \tau_{\theta 1}^i \\ \vdots \\ \tau_{\theta 6}^i \end{bmatrix} = \mathbf{K}_{Foot} \begin{bmatrix} \theta_1^i \\ \vdots \\ \theta_6^i \end{bmatrix}; \quad \begin{bmatrix} \tau_{\theta 7}^i \\ \vdots \\ \tau_{\theta 12}^i \end{bmatrix} = \mathbf{K}_{Leg} \begin{bmatrix} \theta_7^i \\ \vdots \\ \theta_{12}^i \end{bmatrix}, \qquad (6)$$

where $\tau_{\theta j}^i$ is the generalized force for the *j*th virtual joint of the *i*th kinematic chain, $K_{act}$ is the actuator stiffness (scalar), and, $\mathbf{K}_{Foot}$, $\mathbf{K}_{Leg}$ are 6×6 stiffness matrices for the foot and leg respectively. It should be stressed that, in contrast to other works, these matrices are assumed to be non-diagonal. This allows taking into account complicated coupling between rotational and translational deformations, while usual lump-based approach does consider this phenomena [8]. For analytical convenience, expressions (6) may be collected in a single matrix equation

$$\boldsymbol{\tau}_\theta^i = \mathbf{K}_\theta \cdot \delta\boldsymbol{\theta}_i, \quad i = 1,2,3 \qquad (7)$$

where $\boldsymbol{\tau}_\theta^i = (\tau_{\theta 0}^i, \ldots \tau_{\theta 12}^i)^T$ is the aggregated vector of the virtual joint reactions, and $\mathbf{K}_\theta = diag(K_{act}, \mathbf{K}_{Foot}, \mathbf{K}_{Leg})$ is the aggregated spring stiffness matrix of the size 13×13. Similarly, one can define the aggregated vector of the passive joint reactions $\boldsymbol{\tau}_q^i = (\tau_{q1}^i, \ldots \tau_{q4}^i)^T$ but all its components must be equal to zero:

$$\boldsymbol{\tau}_q^i = \mathbf{0}, \quad i = 1,2,3 \qquad (8)$$

To find the static equations corresponding to the end-effector motion $\delta\mathbf{t}_i$, let us apply the principle of virtual work assuming that the joints are given small, arbitrary virtual displacements $(\Delta\boldsymbol{\theta}_i, \Delta\mathbf{q}_i)$ in the equilibrium neighborhood. Then the virtual work of the external force $\mathbf{f}_i$ applied to the end-effector along the corresponding displacement $\Delta\mathbf{t}_i = \mathbf{J}_\theta^i \Delta\boldsymbol{\theta}_i + \mathbf{J}_q^i \Delta\mathbf{q}_i$ is equal to the sum $(\mathbf{f}_i^T \mathbf{J}_\theta^i)\Delta\boldsymbol{\theta}_i + (\mathbf{f}_i^T \mathbf{J}_q^i)\Delta\mathbf{q}_i$. For the internal forces, the virtual work is $-\boldsymbol{\tau}_\theta^{i^T} \cdot \Delta\boldsymbol{\theta}_i$ since the passive joints do not produce the force/torque reactions (the minus sign takes into account the adopted directions for the virtual spring forces/torques). Therefore, because in the static equilibrium the total virtual work is equal to zero for any virtual displacement, the equilibrium conditions may be written as

$$\mathbf{J}_\theta^{i^T} \cdot \mathbf{f}_i = \boldsymbol{\tau}_\theta^i; \qquad \mathbf{J}_q^{i^T} \cdot \mathbf{f}_i = \mathbf{0}. \qquad (9)$$

This gives additional expressions describing the force/torque propagation from the joints to the end-effector. Hence, the complete kinetostatic model consists of five matrix equations (2), (7)…(9) where either $\mathbf{f}_i$ or $\delta\mathbf{t}_i$ are treated as known, and the remaining variables are considered as unknowns. Obviously, since separate kinematic chains posses some degrees-of-freedom, this system cannot be uniquely solved for given $\mathbf{f}_i$. However, vice versa, for given end-effector displacement $\delta\mathbf{t}_i$, it is possible to compute both the corresponding external force $\mathbf{f}_i$ and the internal variables, $\delta\boldsymbol{\theta}_i$, $\boldsymbol{\tau}_\theta^i$, $\delta\mathbf{q}_i$ (i.e. virtual spring reactions and displacements in passive joints, which may also provide useful information for the designer). Since matrix $\mathbf{K}_\theta$ is non-singular (it describes the stiffness of the virtual sprigs), the variable $\delta\boldsymbol{\theta}_i$ can be expressed via $\mathbf{f}_i$ using equations $\boldsymbol{\tau}_\theta^i = \mathbf{K}_\theta \cdot \delta\boldsymbol{\theta}_i$ and $\mathbf{J}_\theta^{i^T} \cdot \mathbf{f}_i = \boldsymbol{\tau}_\theta^i$. This yields substitution $\delta\boldsymbol{\theta}_i = (\mathbf{K}_\theta^{-1}\,\mathbf{J}_\theta^{i^T}) \cdot \mathbf{f}_i$ allowing reducing the kinetostatic model to system of two matrix equations

$$(\mathbf{J}_\theta^i\,\mathbf{K}_\theta^{-1}\,\mathbf{J}_\theta^{i^T}) \cdot \mathbf{f}_i + \mathbf{J}_q^i \cdot \delta\mathbf{q}_i = \delta\mathbf{t}_i; \qquad \mathbf{J}_q^{i^T} \cdot \mathbf{f}_i = \mathbf{0} \qquad (10)$$

with unknowns $\mathbf{f}_i$ and $\Delta\mathbf{q}_i$. This system can be also rewritten in a matrix form

$$\begin{bmatrix} \mathbf{S}_\theta^i & \mathbf{J}_q^i \\ \mathbf{J}_q^{i\,T} & \mathbf{0} \end{bmatrix} \cdot \begin{bmatrix} \mathbf{f}_i \\ \delta\mathbf{q}_i \end{bmatrix} = \begin{bmatrix} \delta\mathbf{t}_i \\ \mathbf{0} \end{bmatrix} \qquad (11)$$

where the sub-matrix $\mathbf{S}_\theta^i = \mathbf{J}_\theta^i\,\mathbf{K}_\theta^{-1}\,\mathbf{J}_\theta^{i^T}$ describes the spring compliance relative to the end-effector, and the sub-matrix $\mathbf{J}_q^i$ takes into account the passive joint influence on the end-effector motions. Therefore, for a separate kinematic chain, the desired stiffness matrix $\mathbf{K}_i$ defining the motion-to-force mapping $\mathbf{f}_i = \mathbf{K}_i \cdot \delta\mathbf{t}_i$, can be computed by direct inversion of relevant 10×10 matrix in the left-hand side of (11) and extracting from it the 6×6 sub-matrix with indices corresponding to $\mathbf{S}_\theta^i$. It is also worth mentioning that computing $\mathbf{S}_\theta^i$ requires 6×6 inversions only, since $\mathbf{K}_\theta^{-1} = diag(K_{act}^{-1}, \mathbf{K}_{Foot}^{-1}, \mathbf{K}_{Leg}^{-1})$. Solvability of system (11) in general case, i.e. for any given $\mathbf{J}_\theta^i$ and $\mathbf{J}_q^i$, cannot be proved. Moreover, if the matrix $\mathbf{J}_q^i$ is singular, the passive joint coordinates $\mathbf{q}_i$ can not be found uniquely. From a physical point of view, it means that if the kinematic chain is located in a singular posture, then certain displacements $\delta\mathbf{t}_i$ can be generated by infinite combinations of the passive joints. But for the variable $\mathbf{f}_i$ the corresponding solution is unique (since the matrix $\mathbf{J}_\theta^i$ is obviously non-singular if at least one 6 d.o.f. spring is included in a serial kinematic chain). On the other hand, the singularity may produce an infinite number of stiffness matrices for the same spatial location of the end-effector and for different values $\mathbf{q}_i$ provided by the inverse kinematics. A special technique to tackle this case, based on the singular value decomposition, has been also developed. After the stiffness matrices $\mathbf{K}_i$ for all kinematic chains are computed, the stiffness of the entire manipulator can be found by simple addition $\mathbf{K}_m = \sum_{i=1}^{3}\mathbf{K}_i$. This follows from the superposition principle, because the total external force corresponding to the end-effector displacement $\delta\mathbf{t}$ (the same for all kinematic chains) can be expressed as $\mathbf{f} = \sum_{i=1}^{3}\mathbf{f}_i$ where $\mathbf{f}_i = \mathbf{K}_i \cdot \delta\mathbf{t}$. It should be stressed that the resulting matrix $\mathbf{K}_i$ is not invertible, since some motions of the end-effector do not produce the virtual spring reactions (because of passive joints influence). However, for the entire manipulator, the stiffness matrix $\mathbf{K}_m$ is s positive definite and invertible for all non-singular (for the rigid model) postures.

### E. Comparison with Other Results

The main advantage of the proposed methodology is its applicability to overconstrained mechanisms. To describe it in more details, let us briefly review an alternative technique [8]. The latter is originated from the same principal equations but the solution strategy includes straightforward elimination of the passive joint variables $\mathbf{q}_i$ using the differential kinematic equations (2) only. Obviously, the feasibility of this step depends on the solvability of the equivalent matrix system

$$\begin{bmatrix} \mathbf{I} & -\mathbf{J}_q^1 & & \\ \mathbf{I} & & -\mathbf{J}_q^2 & \\ \mathbf{I} & & & -\mathbf{J}_q^3 \end{bmatrix} \cdot \begin{bmatrix} \delta\mathbf{t} \\ \delta\mathbf{q}_1 \\ \delta\mathbf{q}_2 \\ \delta\mathbf{q}_3 \end{bmatrix} = \begin{bmatrix} \mathbf{J}_\theta^1 & & \\ & \mathbf{J}_\theta^2 & \\ & & \mathbf{J}_\theta^3 \end{bmatrix} \cdot \begin{bmatrix} \delta\boldsymbol{\theta}_1 \\ \delta\boldsymbol{\theta}_2 \\ \delta\boldsymbol{\theta}_3 \end{bmatrix} \tag{12}$$

where $\delta\mathbf{t}$ and $\delta\mathbf{q}_i$ are treated as unknowns. In the non-constrained case (for the 3-PUU architecture, for instance) the matrix in the left-hand side of (14) is square, of size 18×18, so it can be inverted usually. However, for overconstrained manipulators, this matrix is non-square, so the system cannot be solved uniquely. For example, for manipulators with the parallelogram-type legs (Orthoglide, Delta, etc.) the matrix size is 18×15. So, in [10] three additional (virtual) passive joints were introduced to solve the problem. But, obviously, such a modification changes the manipulator architecture and its stiffness matrix, doubting validity of the corresponding model. Besides, the developed technique allows computing the stiffness matrix even for the singular manipulator postures and does not incorporate the least-square pseudo-inversions applied by other authors. This is achieved by applying another solution strategy, which considers simultaneously the kinematic and static-equilibrium equations for each kinematic chain separately. Some hidden conveniences are included in the modeling stage. In particular, the kinematic models of the chains may include several redundant springs that are totally compensated by relevant passive joints. However, there is no need to eliminate these springs from the model manually, since they do not increase the matrix sizes in system (11). This allows including in the model 6-d.o.f. virtual springs of general type, without any modifications. Another advantage of the proposed technique is that it can be generalized easily. Within this paper, it is applied to the stiffness modeling of 3-d.o.f. translational manipulators with actuators located between the base and the foot. However, it can be easily modified to cover other actuator locations, which may be included in the foot or in the leg. A further generalization is related to a number of kinematic chains and their similarity. They are also not crucial assumptions and influence on the Jacobian computing only. But after the Jacobians are determined, the stiffness matrices for separate chains may be computed in the same manner and then aggregated.

## III. Parameters of the Compliant Elements

The adopted stiffness model of each kinematic chain includes three compliant components, which are described by one 1-d.o.f. spring and two 6-d.o.f. springs corresponding to the actuator, and to the foot/leg links (see Fig. 3). Let us describe particular techniques for their evaluation.

### A. Actuator Compliance

The actuator compliance, described by the scalar parameter $k_{act} = K_{act}^{-1}$, depends on both the servomechanism mechanics and the control algorithms. Since most of modern actuators implement the digital PID control, the main contribution to $k_{act}$ is done by the mechanical transmissions. The latter are usually located outside the feedback-control loop and consist of screws, gears, shafts, belts, etc., whose flexibility is comparable with the flexibility of the manipulator links. Because of the complicated mechanical structure of the servomechanisms, the parameter $k_{act}$ is usually evaluated from static load experiments, by applying the linear regression to the experimental data.

### B. Link Compliance

Following a general methodology, the compliance of a manipulator link (foots and legs) is described by 6×6 symmetrical positive definite matrices $\mathbf{K}_{leg}^{-1}, \mathbf{K}_{foot}^{-1}$ corresponding to 6-d.o.f. springs with relevant coupling between translational and rotational deformations. This distinguishes our approach from other lumped modeling techniques, where the coupling is neglected and only a subset of deformations is taken into account (presented by a set of 1-d.o.f. springs). The simplest way to obtain these matrices is to approximate the link by a beam element for which the non-zero elements of the compliance matrix may be expressed analytically:

$$k_{11} = \frac{L}{EA};\ k_{22} = \frac{L^3}{3EI_z};\ k_{33} = \frac{L^3}{3EI_y};\ k_{44} = \frac{L}{GJ};\ k_{55} = \frac{L}{EI_y};\quad k_{66} = \frac{L}{EI_z};\quad k_{35} = -\frac{L^2}{2EI_y};\quad k_{26} = \frac{L^2}{2EI_z} \tag{13}$$

Here $L$ is the link length, $A$ is its cross-section area, $I_y$, $I_z$, and $J$ are the quadratic and polar moments of inertia of the cross-section, and $E$ and $G$ are the Young's and Coulomb's modules respectively. However, for certain link geometries, the accuracy of a single-beam approximation can be insufficient. In this case the link can be approximated by a serial chain of the beams, whose compliance is evaluated by applying the same method (i.e. considering the kinematic chain with 6-d.o.f. virtual springs, but without passive joints). This leads to the resulting compliance matrix $\mathbf{K}_{Link}^{-1} = \mathbf{J}_b \mathbf{K}_b^{-1} \mathbf{J}_b^T$, where $\mathbf{J}_b$ and $\mathbf{K}_b^{-1}$ incorporate the Jacobian and the compliance matrices for all virtual springs.

### C. FEA-based evaluation of stiffness

For complex link geometries, the most reliable results can be obtained from the FEA modeling. To apply this approach, the CAD model of each link should be extended by introducing an auxiliary 3D object, a "pseudo-rigid" body, which is used as a reference for the compliance evaluation. Besides, the link origin must be fixed relative to the global coordinate system. Then, sequentially and separately applying forces $F_x, F_y, F_z$ and torques $M_x, M_y, M_z$ to the reference object, it is possible to evaluate corresponding linear and angular displacements, which allow computing the stiffness matrix columns. The main difficulty here is to obtain accurate displacement values by using proper FEA-discretization ("mesh size"). Besides, to increase accuracy, the displacements must be evaluated using redundant data set describing the reference body motion. For this reason, it is worth applying a dedicated SVD-based algorithm. As follows from our study, the single-beam approximation of the Orthoglide foot gives accuracy of about 50%, and the four-beam approximation improves it up to 30% only. While the FEA-based method is the most accurate one, it is also the most time consuming. However, in contrast to the straightforward FEA-modeling of the entire manipulator, which requires re-computing for each manipulator posture, the proposed technique involves a single evaluation of link stiffness.

## IV. Application Examples

To demonstrate efficiency of the proposed methodology, let us apply it to the comparative stiffness analysis of two 3-d.o.f. translational mechanism, which employ Orthoglide architecture. CAD models of these mechanisms are presented in Fig. 4.

### A. Stiffness of U-Joint Based Manipulator

First, let us derive the stiffness model for the simplified Orthoglide mechanics, where the legs are comprised of equivalent limbs with U-joints at the ends. Accordingly, to retain major compliance properties, the limb geometry corresponds to the parallelogram bars with doubled cross-section area. Let us assume that the world coordinate system is located at the end-effector reference point corresponding to the isotropic manipulator posture (when the legs are mutually perpendicular and parallel to relevant actuator axes). For this assumption, the geometrical models of separate kinematic chains can be described by the expression (1) Because for the rigid manipulator the end-effector moves with only translational motions, the nominal values of the passive joint coordinates are subject to the specific constraints $q_3 = -q_2$; $q_4 = -q_1$, which are implicitly incorporated in the direct/inverse kinematics [10]. However, the flexible model allows variations for all passive joints. Using the link stiffness parameters obtained by the FEA-modeling and applying the proposed methodology, we computed the compliance matrices for three typical manipulator postures, the principal components of which are presented in Table 1. Below, they are compared with the compliance of the parallelogram-based manipulator.

### B. Stiffness of Parallelogram Based Manipulator

Before evaluation the compliance of the entire manipulator, let us derive the stiffness matrix of the parallelogram. Using the adopted notations, the parallelogram equivalent model may be written as

$$\mathbf{T}_{Plg} = \mathbf{R}_y(q_2) \cdot \mathbf{T}_x(L) \cdot \mathbf{R}_y(-q_2) \cdot \mathbf{V}_s(\theta_7, \ldots \theta_{12}) \tag{14}$$

where, compared to the above case, the third passive joint is eliminated (it is implicitly assumed that $q_3 = -q_2$). On the other hand, the original parallelogram may be split into two serial kinematic chains (the "upper" and "lower" ones)

$$\begin{aligned}\mathbf{T}_{up} = {} & \mathbf{T}_z(-d/2) \cdot \mathbf{R}_y(q + \Delta q_1^{up}) \cdot \mathbf{T}_x(L) \cdot \\ & \cdot \mathbf{V}_s(\theta_1^{up}, \ldots \theta_6^{up}) \cdot \mathbf{R}_y(-q + \Delta q_2^{up}) \cdot \mathbf{T}_z(d/2)\end{aligned} \tag{15}$$

$$\begin{aligned}\mathbf{T}_{dn} = {} & \mathbf{T}_z(d/2) \cdot \mathbf{R}_y(q + \Delta q_1^{dn}) \cdot \mathbf{T}_x(L) \cdot \\ & \cdot \mathbf{V}_s(\theta_1^{dn}, \ldots \theta_6^{dn}) \cdot \mathbf{R}_y(-q + \Delta q_2^{dn}) \cdot \mathbf{T}_z(-d/2)\end{aligned} \tag{16}$$

where $L$, $d$ are the parallelogram geometrical parameters, $\Delta q_1^i, \Delta q_2^i$, $i \in \{up, dn\}$ are the variations of the passive joint coordinates and the sub/superscripts "*up*" and "*dn*" correspond to the upper and lower chain respectively. Hence, the parallelogram compliance matrix may be also derived using the proposed technique that yields an analytical expression

$$\mathbf{K}_{Plg} = 2 \cdot \left[\begin{array}{ccc|ccc} K_{11} & 0 & 0 & 0 & 0 & 0 \\ 0 & K_{22} & 0 & 0 & 0 & K_{26} \\ 0 & 0 & 0 & 0 & 0 & 0 \\ \hline 0 & 0 & 0 & K_{44} + \frac{d^2 C_q^2 K_{22}}{4} & 0 & \frac{d^2 S_{2q} K_{22}}{8} \\ 0 & 0 & 0 & 0 & \frac{d^2 C_q^2 K_{11}}{4} & 0 \\ 0 & K_{26} & 0 & \frac{d^2 S_{2q} K_{22}}{8} & 0 & K_{66} + \frac{d^2 S_q^2 K_{22}}{4} \end{array}\right] \tag{17}$$

where $C_q = \cos(q)$; $S_q = \sin(q)$. Using this model and applying the proposed technique, we computed the compliance matrices for three typical manipulator postures (see table Table 1). As follows from the comparison with the U-joint case, the parallelograms allow increasing the rotational stiffness roughly in 10 times. This justifies application of this architecture in the Orthoglide prototype design [15].

## V. Conclusions

The paper proposes a new systematic method for computing the stiffness matrix of overconstrained parallel manipulators. It is based on multidimensional lumped model of the flexible links, whose parameters are evaluated via the FEA modeling and describe both the translational/rotational compliances and the coupling between them. In contrast to previous works, the method employs a new solution strategy of the kinetostatic equations, which considers simultaneously the kinematic and static relations for each separate kinematic chain and then aggregates the partial solutions in a total one. This allows computing the stiffness matrices for overconstrained mechanisms for any given manipulator posture, including singular configurations and their neighborhood. Another advantage is computational simplicity that requires low-dimensional matrix inversion compared to other techniques. Besides, the method does not require manual elimination of the redundant spring corresponding to the passive joints, since this operation is inherently included in the numerical algorithm. The efficiency of the proposed method was demonstrated through application examples, which deal with comparative stiffness analysis of two parallel manipulators of the Orthoglide family (with U-joint based and parallelogram based links). Relevant simulation results have confirmed essential advantages of the parallelogram based architecture and validated adopted design of the Orthoglide prototype. Another contribution is the analytical stiffness model of the parallelogram, which was derived using the same methodology. While applied to the 3-d.o.f. translational mechanisms, the method can be extended to other parallel architectures composed of several kinematic chains with rotational/prismatic joints and limb- or parallelogram-based links. So, future work will focus on the stiffness modeling of more complicated parallel mechanism with another actuator location (such as the Verne machine [16]) and also on the experimental verification of the stiffness models for the Orthoglide robot.

## REFERENCES

[1] J. Tlusty, J. Ziegert and S. Ridgeway, "Fundamental Comparison of the Use of Serial and Parallel Kinematics for Machine Tools," In: Annals of the CIRP, vol. 48(1), 1999.

[2] P. Wenger, C. M. Gosselin and B. Maillé, "A Comparative Study of Serial and Parallel," In: Mechanism Topologies for Machine Tools, PKM'99, pp. 23-32, Milano, 1999.

[3] F. Majou, P. Wenger and D. Chablat, "The design of Parallel Kinematic Machine Tools using Kinetostatic Performance Criteria," In: 3rd International Conference on Metal Cutting and High Speed Machining, Metz, France, June 2001.

[4] G. Pritschow and K.-H. Wurst, "Systematic Design of Hexapods and Other Parallel Link Systems," In: Annals of the CIRP, vol. 46(1), pp 291–295, 1997.

[5] O. Company and F. Pierrot, "Modelling and Design Issues of a 3-axis Parallel Machine-Tool," Mechanism and Machine Theory, vol. 37, pp. 1325–1345, 2002.

[6] T. Brogardh, "PKM Research - Important Issues, as seen from a Product Development Perspective at ABB Robotics," In: Workshop on Fundamental Issues and Future Research Directions for Parallel Mechanisms and Manipulators, Quebec, Canada, October 2002.

[7] A. Paskhevich, P. Wenger and D. Chablat, "Kinematic and stiffness analysis of the Orthoglide, a PKM with simple, regular workspace and homogeneous performances," In: IEEE International Conference On Robotics And Automation, Rome, Italy, April 2007

[8] C.M. Gosselin, "Stiffness mapping for parallel manipulators," IEEE Transactions on Robotics and Automation, vol. 6, pp. 377–382, 1990.

[9] X. Kong and C. M. Gosselin, "Kinematics and Singularity Analysis of a Novel Type of 3-CRR 3-DOF Translational Parallel Manipulator," The International Journal of Robotics Research, vol. 21(9), pp. 791–798, September 2002.

[10] F. Majou, C. Gosselin, P. Wenger and D. Chablat. "Parametric stiffness analysis of the Orthoglide," Mechanism and Machine Theory, vol. 42(3), pp. 296–311, March 2007.

[11] B.C. Bouzgarrou, J.C. Fauroux, G. Gogu and Y. Heerah, "Rigidity analysis of T3R1 parallel robot uncoupled kinematics," In: Proc. of the 35th International Symposium on Robotics (ISR), Paris, France, March 2004.

[12] D. Deblaise, X. Hernot and P. Maurine, "A Systematic Analytical Method for PKM Stiffness Matrix Calculation," In: IEEE International Conference on Robotics and Automation (ICRA), pp. 4213–4219, Orlando, Florida, May 2006.

[13] Y. Li and Q. Xu, "Stiffness Analysis for a 3-PUU Parallel Kinematic Machine", Mechanism and Machine Theory, (In press: Available online 6 April 2007)

[14] R. Clavel, "DELTA, a fast robot with parallel geometry," Proceedings, of the 18th International Symposium of Robotic Manipulators, IFR Publication, pp. 91–100, 1988.

[15] D. Chablat and Ph. Wenger, "Architecture Optimization of a 3-DOF Parallel Mechanism for Machining Applications, the Orthoglide," IEEE Transactions On Robotics and Automation, vol. 19(3), pp. 403–410, 2003.

[16] D. Kanaan, P. Wenger and D. Chablat, "Kinematics analysis of the parallel module of the VERNE machine," In: 12th World Congress in Mechanism and Machine Science, IFToMM, Besançon, June 2007.

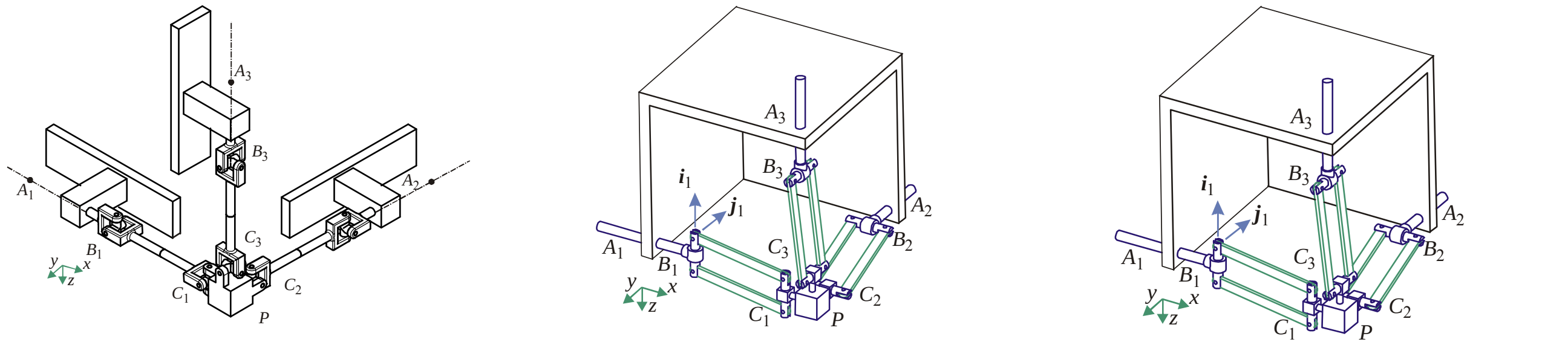

(A) U-JOINT BASED ARCHITECTURE (B) PARALLELOGRAM BASED ARCHITECTURE (C) WORKSPACE AND CRITICAL POINTS $Q_1$ AND $Q_2$

FIG. 4. KINEMATICS OF TWO 3-DOF TRANSLATIONAL MECHANISMS EMPLOYING THE ORTHOGLIDE ARCHITECTURE

TABLE I: TRANSLATIONAL AND ROTATIONAL STIFFNESS OF THE 3-PUU AND 3-PRPAR MANIPULATORS

| MANIPULATOR ARCHITECTURE | Point $Q_0$ $x, y, z = 0.00\ mm$ | | Point $Q_1$ $x, y, z = -73.65\ mm$ | | Point $Q_2$ $x, y, z = +126.35\ mm$ | |
|---|---|---|---|---|---|---|
| | $k_{tran}$ [N/mm] | $k_{rot}$ [N·mm/rad] | $k_{tran}$ [N/mm] | $k_{rot}$ [N·mm/rad] | $k_{tran}$ [N/mm] | $k_{rot}$ [N·mm/rad] |
| 3-PUU manipulator | $2.78 \cdot 10^{-4}$ | $20.9 \cdot 10^{-7}$ | $10.9 \cdot 10^{-4}$ | $24.1 \cdot 10^{-7}$ | $71.3 \cdot 10^{-4}$ | $25.8 \cdot 10^{-7}$ |
| 3-PRPaR manipulator | $2.78 \cdot 10^{-4}$ | $1.94 \cdot 10^{-7}$ | $9.86 \cdot 10^{-4}$ | $2.06 \cdot 10^{-7}$ | $21.2 \cdot 10^{-4}$ | $2.65 \cdot 10^{-7}$ |